%% file: content_analysis.tex
\documentclass{article}
\usepackage{spconf}

\usepackage{amsmath,amssymb,amsfonts}
\usepackage{bbm}
\usepackage{epsfig}
\usepackage{cite}
\usepackage{algorithmic}
\usepackage{graphicx}
\usepackage{textcomp}
\usepackage{float}
\usepackage{xcolor}
\usepackage{threeparttable}
\usepackage{tabularx}
\usepackage{multirow}
\usepackage[caption=false,font=footnotesize]{subfig}
\usepackage{soul}

\ifodd 1
\newcommand{\CommentWong}[1]{\textcolor[rgb]{0,0.3,1}{[Wong: #1]}}

\newcommand{\CommentZhao}[1]{\textcolor[rgb]{1,0,0}{[Zhao: #1]}}

\else
\newcommand{\CommentWong}[1]{}

\newcommand{\CommentZhao}[1]{}

\fi

\graphicspath{{figures/}}

\newcommand{\MyParagraph}[1]{\vspace{1mm}\noindent{}\textbf{#1}\hspace{3mm}}

\def\anger{\textsc{Anger}}
\def\disappoint{\textsc{Disappoint}}
\def\sorrow{\textsc{Sorrow}}
\def\fear{\textsc{Fear}}
\def\worry{\textsc{Worry}}
\def\satisfied{\textsc{Satisfied}} % /Happiness
\def\happiness{\textsc{Satisfied/Happiness}}
\def\hope{\textsc{Hope}}
\def\empathy{\textsc{Empathy/Sympathy}}

\def\grateful{\textsc{Grateful}}
\def\surprise{\textsc{Surprise}}
\def\sarcasm{\textsc{Sarcasm}}
\def\na{\textsc{Na}}

\begin{document}
\ninept

%\sloppy

\title{Toward Effective Automated Content Analysis via Crowdsourcing\vspace{-1mm}}

\name{Jiele Wu$^{1,2}$, 
Chau-Wai Wong$^2$, 
Xinyan Zhao$^3$,
Xianpeng Liu$^2$
\thanks{Part of the work was completed when Xinyan Zhao was with the Hong Kong Baptist University (HKBU). Data collection was supported by a seed grant from the school of communication at the HKBU.}
\vspace{-2mm}}

\address{$^1$\,Beijing Institute of Technology, China\hspace{8mm}
$^2$\,North Carolina State University, USA\hspace{0mm}\\
$^3$\,University of North Carolina at Chapel Hill, USA
\vspace{-3mm}}

\maketitle

\begin{abstract}%
Many computer scientists use the aggregated answers of online workers to represent ground truth. Prior work has shown that aggregation methods such as majority voting are effective for measuring relatively objective features. For subjective features such as semantic connotation, online workers, known for optimizing their hourly earnings, tend to deteriorate in the quality of their responses as they work longer. In this paper, we aim to address this issue by proposing a quality-aware semantic data annotation system. We observe that with timely feedback on workers' performance quantified by quality scores, better informed online workers can maintain the quality of their labeling throughout an extended period of time. We validate the effectiveness of the proposed annotation system through i) evaluating performance based on an expert-labeled dataset, and ii) demonstrating machine learning tasks that can lead to consistent learning behavior with 70\%--80\% accuracy. Our results suggest that with our system, researchers can collect high-quality answers of subjective semantic features at a large scale.
\end{abstract}


\input{1_intro}

\input{2_related_work}

\input{3_system_design}

\input{4_majority_vote}

\input{5_discuss}

\vspace{-2mm}
\bibliographystyle{IEEEtran}
\bibliography{content_analysis}

\end{document}

%% file: 1_intro.tex
\section{Introduction}

Content analysis is a well-established social scientific methodology that can provide high-quality labels especially for subjective features\cite{riffe2014}. Standard procedures of content analysis involve domain theory-based protocol development and intensive coder training spanning through weeks or months\cite{krippendorff-2004}. Due to the time-consuming nature, it has been difficult to apply content analysis to big data research.
Crowdsourcing has been used to solve a wide range of labeling tasks and is the most prevalent method of dataset collection in computer science\cite{datacollectionsurvey}. Social scientists have also begun to investigate the applicability of crowdsourcing for generating quality labels for subjective features. Existing studies show that  crowdsourcing is useful for simpler coding/annotation tasks\cite{Lind-2017}, whereas challenges remain for using crowdsourcing in complex coding tasks involving subjective semantic features such as frames\cite{Guo-2020}.

There are mainly two approaches to data annotation.
First, with the computer science approach, researchers recruit a large pool of online workers from Amazon Mechanical Turk (MTurk) or other platforms to obtain data labels. 
To optimize the monetary gain, online workers may prioritize quantity over quality in coding and their performance may decrease over time. 
Because online workers are not likely to be penalized
for their low-quality answers, the quality of the collected labels, particularly for subjective features, cannot be ensured \cite{Song-2020}.  Second, with the social science approach, researchers typically spend an extensive amount of time developing domain theory-informed protocol and training a small number of expert coders to provide accurate and consistent answers. The quality of labeling is usually guaranteed by ensuring inter-coder agreement \cite{krippendorff-2004}. Yet, the dataset built through this approach is neither scalable nor efficient.  

Our goal is to propose a system that can harvest a large number of high-quality labels via crowdsourcing platforms while ensuring the quality of labels collected over time. Such a system can be particularly useful for complex coding tasks that involve the labeling of subjective or latent semantic features. 
The contributions of this paper are as follows.
First, we propose a quality-aware annotation system for collecting a large number of high-quality labels from online workers. Such a system can be particularly useful for complex coding tasks involving big data.
We also provide recommendations of experimental design when the system is deployed for emotion label collection.
Second, we show that labels aggregated based on majority voting through our system can achieve accuracy, supported by a comparison of crowd-labeled data and expert-labeled data.
Third, our results reveal that majority voting and other proposed weighted voting schemes can produce labels that are consistent and learnable.

%% file: 2_related_work.tex
\section{Related Work}

\MyParagraph{Content Analysis}
Content analysis, as a social science research tool, is the quantitative, objective, and systematic examination of content and the analysis of relationships involving these values based on statistical methods\cite{riffe2014}. Traditionally, social scientists heavily rely on human coding to measure subjective or latent features. Latent features involve semantic implications that can vary among individuals \cite{potter-1999}, such as sentiment, discrete emotions, or media frames. Content analysis can help a social scientist to collect valid and replicable labels. Validity is the extent to which a latent feature is actually measured through content analysis \cite{riffe2014}. Reliability is the extent to which collected labels through the same coding rules are consistent over coders, location, and time \cite{krippendorff-2004}.

In recent decades, automated content analysis has been prospering, with the burgeoning application of dictionary-based tools such as Linguistic Inquiry and Word Count (LIWC)\cite{Tausczik-2010} and machine learning algorithms among social scientists in various fields.  While the performance of existing automated content analysis methods is relatively satisfactory in capturing certain features such as sentiments\cite{Tausczik-2010}, there still lacks validated tools to accurately capture more subjective or latent features. A large amount of human coding is still much needed for evaluating computer models for addressing the concern of validity\cite{Guo-2020}. 
    
\MyParagraph{Crowdsourcing for Complex Coding Tasks}
Crowdsourcing can be used to outsource human coding tasks to a large group of human workers online. In computer science, crowdsourcing has been widely used to train and evaluate prediction models. Recently, social scientists have begun to explore the applicability of crowdsourcing to provide ground truth for annotation. Their results show that crowdsourcing can provide quality labels for less complex coding tasks such as sentiment analysis\cite{Lind-2017}. Yet, crowdsourcing may be less applicable for more complex tasks such as frame analysis or moral judgment\cite{Weber-2018}. Recent work also suggests the importance of quality control of human coders' performance by showing that human judgment, subject to errors, can deviate from the golden standard \cite{Song-2020}.

%% file: 3_system_design.tex
\section{Quality-Aware Annotation System}
\label{sec:system_design}
In this section, we explain the design of the proposed quality-aware semantic annotation system through a task of labeling the emotions of tweets related to the Flint water crisis.

\vspace{1mm}\noindent{}\textbf{Motivations}\hspace{3mm} 
Using crowdsourcing platforms such as Amazon Mechanical Turk (MTurk) for complex coding tasks is more challenging than conducting traditional content analysis.
First, a qualifying process is necessary for selecting MTurk workers who are more capable of completing complex coding tasks. 
Our task of tweet emotion labeling involves more than ten emotions under three major categories.
It requires workers to be able to precisely understand the definition of each emotion and to differentiate emotions of subtle differences.
However, MTurk is dominated by easy tasks that require light mental efforts, such as receipts transcription, visual object classification, and information verification via a Wikipedia page.
Second, the worker--requester dynamics on MTurk are different from the coder--researcher dynamics in a traditional content analysis setup.
Most MTurk workers are veterans in optimizing hourly earnings, and the quality of their responses tends to deteriorate as they work longer if no quality control is in place according to our observation.
This is different from a traditional content analysis setup in which coders are not motivated to game the system. 
We therefore would like to have the capability to closely monitor workers' performance as they keep working on the assignments.

\vspace{1mm}\noindent{}\textbf{System Overview}\hspace{3mm} 
Our system collects annotated labels from online workers recruited through and work for the MTurk service.
A candidate worker needs to go through a qualifying process and will only be allowed to work on our annotation tasks if his/her score passes a baseline.
The annotation tasks are administered via the MTurk assignments, each of which contains 20 unlabeled tweets randomly selected from a pool of 9,287 tweets that we pre-selected based on the existence of emotion-related words.
Once the worker submits a completed assignment, it will be graded by the system based on a small portion of questions with ground-truth labels that were randomly embedded into the assignment.
The worker will receive a quality score as feedback for the current assignment and a cumulative quality score measuring his/her overall performance.
A worker must maintain a cumulative quality score no less than a certain threshold in order to be qualified for further tasks.
The system will record other metadata generated during the process of answering the assignments to facilitate further data mining.

\vspace{1mm}\noindent{}\textbf{Qualifying Process and Real-Time Performance Monitoring}\hspace{3mm}
Prior research\cite{Weber-2018} has shown that it is challenging to collect high-quality semantic labels for complex coding tasks.
Our system is designed to address this challenge by a qualifying process that selects capable workers and by a real-time performance monitoring scheme to maintain the quality of collected labels.

The \emph{qualifying process} is implemented by a paid qualification task consisting of a training session and a test session.
The training session contains the background of the task, definitions of emotions with example tweets, labeling instructions with hints, and five multiple-choice training questions.
A worker must correctly answer a multiple-choice question before being given the next question.
The worker can validate the answers by clicking a ``verify'' button and hints will be shown if the answer is incorrect.
Once the worker manages to complete the training session, he/she will be asked to complete the test session comprising another 15 questions for assessing his/her capability.
If the worker scores no less than 40\%, he/she will be considered qualified and formally enrolled as a worker for our annotation tasks.

The \emph{real-time performance monitoring} will measure workers' performance and exclude lower performing workers.
Two quality scores are considered, namely, the assignment quality score and the cumulative quality score.
The \emph{assignment quality score} (or simply the \emph{quality score}, when the context is clear) reflects how well a worker answers an assignment.
Five randomly selected questions with the ground-truth labels were mixed into 20 questions.
The assignment quality score is estimated by the percentage of correctly answered questions out of the five embedded ones. 
This estimator is unbiased but noisy.
The \emph{cumulative quality score} measuring a particular worker's performance is defined as the sample mean of his/her assignment quality scores to date. 
This aggregated score is less noisy than the assignment score due to the central limit theorem, especially for those workers who have completed a large number of assignments.
In our experiment, we require a worker to keep a cumulative score above \emph{minimally qualifying score}, e.g., 60\%, to be qualified for further tasks.
If a worker completes an assignment with a quality score no greater than 40\%, he/she will be issued a warning message.
This design keeps workers informed of their averaged performance and can help to maintain the overall quality of the annotated data.

\vspace{1mm}\noindent{}\textbf{Ground-Truth Data Labeling}\hspace{3mm}
One hundred tweets labeled by expert coders through standard content analytical procedures were used as the ground truth to assess workers' performance. Identifying the true emotions for tweets requires heavy mental efforts, relevant background knowledge, and is thus highly nontrivial. 
Two social scientists  with expertise in the subject matter were asked to provide labels that approximate the ground truth. They followed procedures in content analysis\cite{riffe2014} to ensure the validity and reliability of coding. First, they created a protocol that records domain theory-based coding rules for each emotion. They then went through several rounds of training. In each round, they independently labeled a small set of tweets and discussed their discrepancy in coding. In this process, they fine-tuned the coding rules in the protocol and reached a consensus on the emotion labels.

\vspace{1mm}\noindent{}\textbf{Emotion Category Creation and Questionnaire Preparation}\hspace{3mm}
We collected a complete set of tweets using keywords related to the Flint water crisis, and used a sentiment analysis tool named VADER\cite{Hutto-2014} to identify those tweets with emotion-related words for our study. To decide which emotion types to focus on, we followed the psychology and communication literature \cite{Lazarus-1991, Jin-2012} to code the following emotion categories: 
i) negative emotions, including \anger, \disappoint, \sorrow, \fear, and \worry, 
ii) positive emotions, including \satisfied, \hope, \empathy, and \grateful, and 
iii) neutral emotions, including \surprise{} and \sarcasm. Not applicable (\na) means that a tweet expresses no emotion.
Based on the recommendation of Benoit et al.\cite{benoit-2016}, we aimed to have each tweet labeled five times for a total of 9,287 tweets. 
We pre-generated 2,817 assignments and stored them in the system for later use.
We collected a total of 42,980 labels, with almost all tweets being labeled five times.
The count of each category is summarized in Table~\ref{tab:label_stat}.

\begin{table}[!t]
\centering
\vspace{-2mm}
\caption{Distribution of the Number of Annotated Emotions for the Flint Water Crisis Dataset\vspace{1mm}}
\label{tab:label_stat}
\scalebox{0.9}{
\begin{tabular}{lrcc}
\hline\hline
\textbf{Emotion} & \textbf{Count} & \textbf{Valence} & \textbf{Count}\\
\hline  
Anger  & 2,550 & \multirow{5}{*}{Negative} & \multirow{5}{*}{5,520}\\
Disappoint  & 1,337 & &\\
Sorrow  & 1,168 & &\\
Fear  & 204 & & \\
Worry  & 261 & &\\
\hline
Happiness/satisfied  & 699 & \multirow{4}{*}{Positive} & \multirow{4}{*}{2,531}\\
Hope  & 675 &  & \\
Empathy/sympathy  & 305 & &\\
Grateful  & 852 & &\\
\hline  
Surprise  & 436 & \multirow{2}{*}{Neutral} & \multirow{2}{*}{1,161}\\
Sarcasm  & 725 & &\\
\hline
N/A  & 75 & &\\
\hline\hline
\end{tabular}
}
\end{table}

%% file: 4_majority_vote.tex
%\section{Majority-Voted Labels are Consistent and Learnable}
\section{Proposed Annotation System is Effective}
\label{sec:ml}

In this section, we will verify via data analysis and machine learning the effectiveness of the proposed quality-aware semantic data annotation system.
In Section~\ref{subsec:quality_control}, we will show that only a small fraction of MTurk workers are willing or able to deliver at a required level of accuracy the responses of complex coding tasks.
In Section~\ref{subsec:validity}, we will verify that majority voting by qualified workers is consistent with expert labeling.
In Section~\ref{subsec:neural_net}, we will verify the learnability of the collected data via experiments that the majority-voting labels can be predicted from input tweets by a state-of-the-art learning system.

\subsection{Quality Control is a Must for Complex Coding Tasks}
\label{subsec:quality_control}

Given the dynamics on MTurk and our requirement for complex coding task, we implemented a paid qualifying process to recruit MTurk workers that are both willing and able to answer complex coding questions.
Our qualifying task attracted 1,030 workers, but only 150 workers managed to pass the training session and successfully submitted the task.
This shows that only a small fraction of MTurk workers are interested in and capable of doing our complex coding task.
Should the qualification test not be in place, we expect to collect most results from uncommitted workers, which can potentially lower the quality of the collected data.

Once qualified, 87 workers, who submitted two or more assignments, form the pool of workers who contributed to the labeling tasks.
Workers were reminded about their cumulative quality scores upon submissions and the minimally qualifying score, which is a designed mechanism to encourage better label quality while balancing out the negative impact due to the temptation of the workers to maximizing their hourly earnings.
Under such dynamics, 10 workers or 11.5\% of the total workers were not able to maintain their cumulative quality scores above the minimally qualifying score and were subsequently disqualified from the task. 
This indicates that the real-time performance monitoring mechanism is effective in removing weak workers for the complex coding task.

\subsection{Majority Voting is Consistent with Experts Labeling}
\label{subsec:validity}

The main purpose of this experiment is to examine the relationship between expert-based scores and majority-vote-based scores.
For each submitted assignment, we associate it with two scores, namely, i) the expert-based score calculated from the five embedded tweets that have ground-truth labels, and ii) the majority-voting-based score for the 20 remaining questions.
Note that when the majority voting results are used to assess the 20 questions, they are assumed to be ``correct.'' 
We collect the score pairs for all submitted assignments for all workers to evaluate the consistency of the pair.
We use Pearson's correlation coefficient $\rho$ and the mean absolute error (MAE) as the evaluation metrics.
Given two vectors ${\bf u}_1 = (u_{1i}, \dots, u_{1n})$ and ${\bf u}_2 = (u_{2i}, \dots, u_{2n})$, the correlation coefficient and MAE are defined as follows:
\begin{equation}
    \rho({\bf u}_1,{\bf u}_2) = \cfrac{\sum_{i=1}^n (u_{1i}-\bar{u}_1)(u_{2i}-\bar{u}_2)}{\sqrt{\sum_{i=1}^n (u_{1i}-\bar{u}_1)^2 \sum_{i=1}^n (u_{2i}-\bar{u}_2)^2}},
\end{equation}
\begin{equation}
    \operatorname{MAE}({\bf u}_1,{\bf u}_2) = \cfrac{1}{n}\sum_{i=1}^n |u_{1i} - u_{2i}|,
\end{equation}
where $\bar{u}_k = \frac{1}{n} \sum_{i=1}^n u_{ki}$ is the sample mean.

\begin{figure}[!t]
  \centering
  \vspace{-5mm}
  \subfloat[][]{\includegraphics[width=0.495\linewidth, trim=0cm 0cm 24mm 0cm, clip]{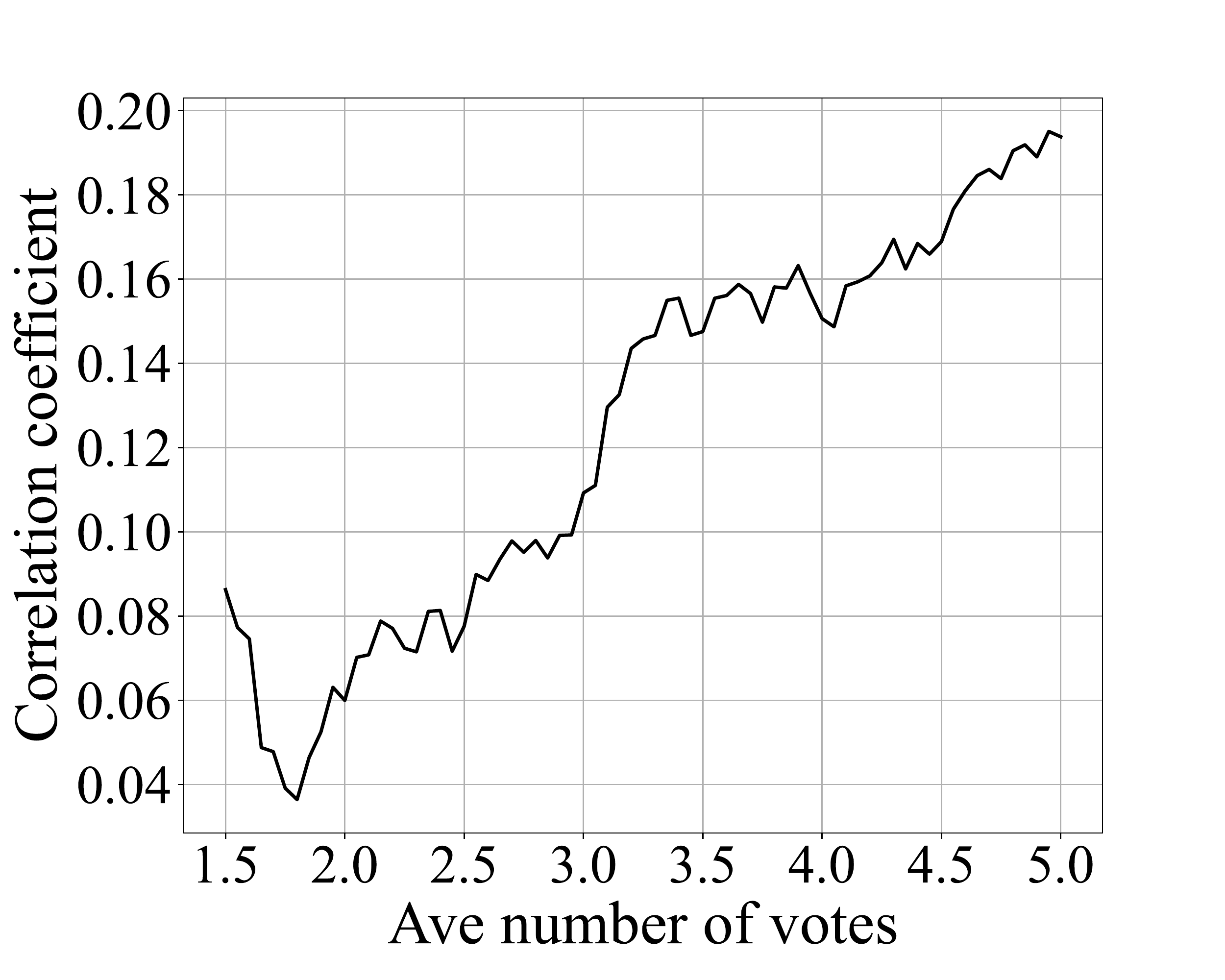}}
  \subfloat[][]{\includegraphics[width=0.495\linewidth, trim=0cm 0cm 24mm 0cm, clip]{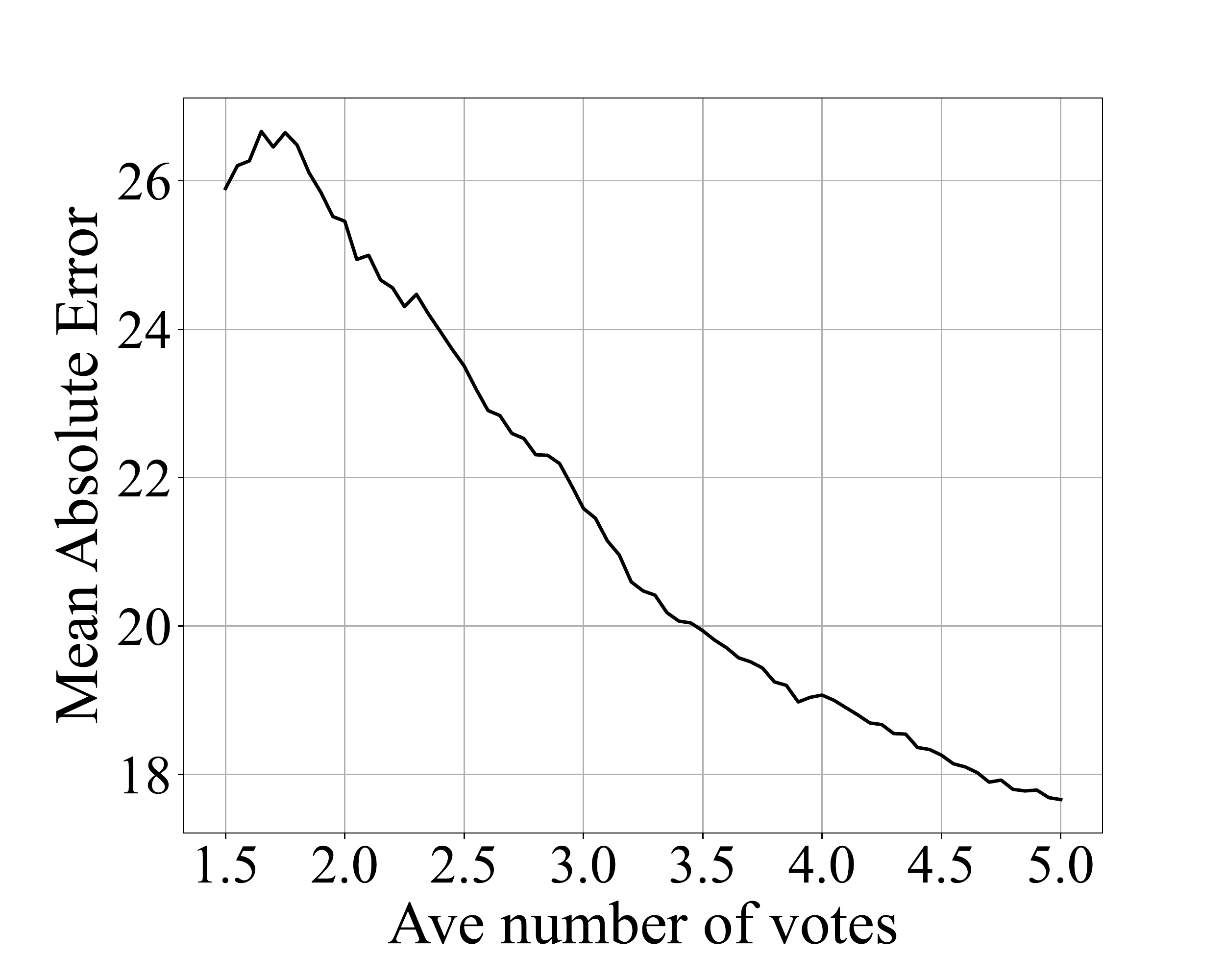}}\\
  \vspace{-4mm}
  \subfloat[][]{\includegraphics[width=0.48\linewidth, trim=0cm 0cm 24mm 0cm, clip]{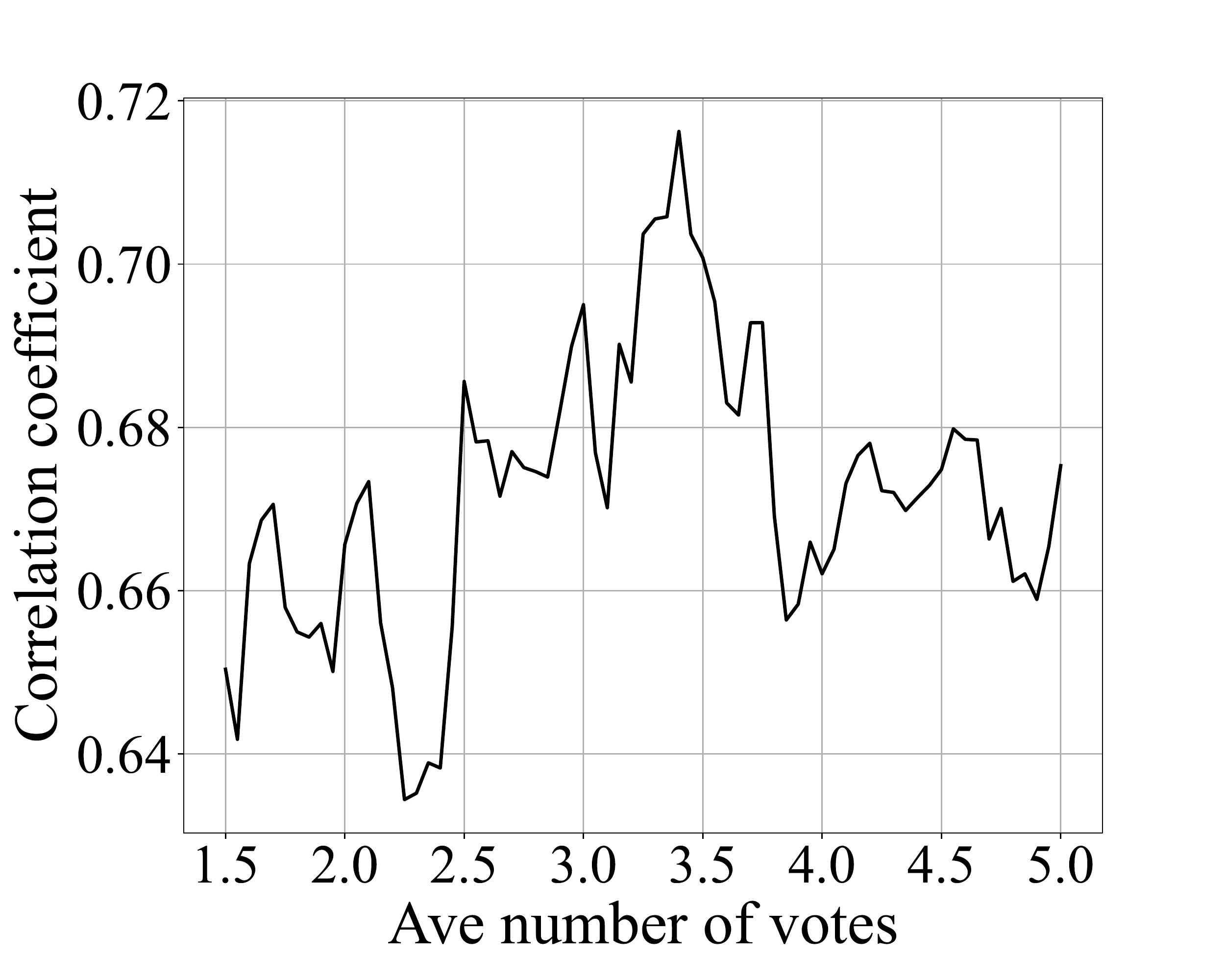}}
  \hspace{1mm}
  \subfloat[][]{\includegraphics[width=0.48\linewidth, trim=0cm 0cm 24mm 0cm, clip]{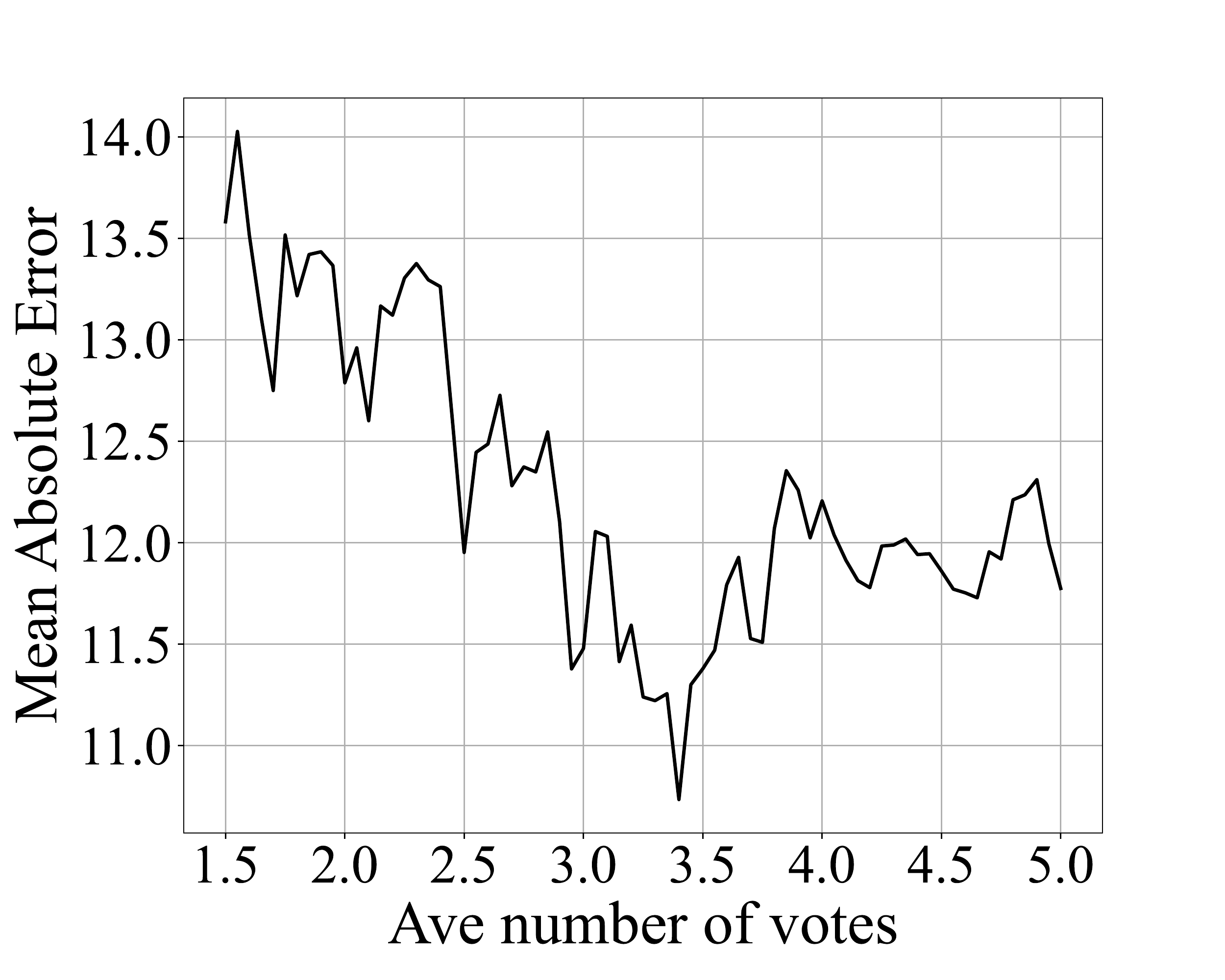}}
  \vspace{-2mm}
  \caption{Consistency between the assignment quality score and the majority-voting quality score. First row: major-voting quality scores evaluated on 20 tweets to be labeled. The improving trend implies that the results of majority voting can be more trustworthy as the number of votes increases. Second row: major-voting quality scores evaluated on 5 profiling tweets. The correlation between the majority voting and expert labeling is about 0.67.
  }
	\vspace{-2mm}
  \label{fig:performance_when_more_data_used}
\end{figure}

We are interested in examining the trend of consistency as more votes are used for majority voting.
Our semantic annotation system is configured to collect 5 votes for each tweet, and at any time before the completion of data collection, the average number of votes is less than 5.
We can therefore gradually increase the proportion of collected votes, $\eta \in [0, 1]$, based on the timestamps of submissions and examine how the consistency evolves.
We exclude workers who contributed less than 5 assignments to avoid potentially contaminating the dataset by abnormal workers.
We calculate the correlation and MAE and plot them against the average number of votes, $5\eta$, in Figs.~\ref{fig:performance_when_more_data_used}(a) and (b), respectively.
The curves reveal that the consistency in terms of correlation and MAE has an improving trend as the averaged number of votes used for majority voting increases from 1.5 votes to 5 votes.

We also use five embedded tweets to evaluate the consistency between the majority voting and expert labeling.
Fig.~\ref{fig:performance_when_more_data_used}(c) reveals a significantly increased correlation ($\sim$0.67) compared to Fig.~\ref{fig:performance_when_more_data_used}(a), as correlations in Fig.~\ref{fig:performance_when_more_data_used}(c) are  calculated from the same set of embedded tweets.
The correlation curve in Fig.~\ref{fig:performance_when_more_data_used}(c) is noisier than that of Fig.~\ref{fig:performance_when_more_data_used}(a) after switching from using 20 tweets drawn from a pool of 9,287 tweets to 5 embedded tweets drawn from a pool of 100 tweets for calculating the majority-voting-based score.
In sum, the two correlation analyses reveal that, for this complex coding task, majority voting has achieved meaningful agreement with expert labeling.

\subsection{Majority Voting Results Are Learnable}
\label{subsec:neural_net}

In this subsection, we investigate the learnability of the labels collected by our proposed quality-aware semantic annotation system.
We first reason that the learnability can be characterized using the generalization capability of a powerful learning system.
We then show that majority-voting based labels can be learned by a deep neural network, achieving a classification accuracy of 70\%--80\%.

\MyParagraph{Definition of Learnability}
We reason that the generalization capability\cite{ESL-ch7} of a powerful learning system on a dataset can quantify the learnability of the dataset.
We define a (supervised machine) \emph{learning system} in the context of our tweet--emotion dataset as follows.
Given a training set that consists of input tweets and their corresponding output emotion labels, a learning system is capable of relating these inputs and outputs in an insightful way such that the learned system is capable of predicting the output labels using an unseen input from a test set.
The accuracy measured on the test set is defined to be the generalization capability.
Here, it is well understood that the training set and test set are from the same population and the learning system has no access to the test set during training.
We further define \emph{a powerful learning system} as the best performing learning system of a set of candidate learning systems that devote their best efforts to related the output labels to input tweets.
While in the traditional machine learning context, the generalization capability is a characteristic of a learning system, from the perspective of a given dataset, \emph{the generalization capability of the powerful learning system} can be considered as how easy the dataset can be consistently learned by this powerful learner. We therefore name it as the \emph{learnability}.

\begin{table}[!t]
\vspace{-2mm}
\caption{Learnability of Labels by Majority Voting and Other Weighted Voting Schemes on Three Different Datasets\vspace{1mm}}
\label{tab:nn_results}
\centering
\scalebox{0.96}{
\begin{tabular}{l|cc|cc|cc}
\hline\hline
\multicolumn{1}{c|}{\multirow{3}{*}{\textbf{\begin{tabular}[c]{@{}c@{}}Weight\\ scheme\end{tabular}}}} & \multicolumn{2}{c|}{\textbf{Valence}} & \multicolumn{2}{c|}{\textbf{Resiliency}} & \multicolumn{2}{c}{\textbf{Attribution}} \\ \cline{2-7} 
\multicolumn{1}{c|}{} 
& \multicolumn{2}{c|}{\textbf{Balanced acc}} 
& \multicolumn{2}{c|}{\textbf{Balanced acc}}
& \multicolumn{2}{c}{\textbf{Balanced acc}} \\ \cline{2-7} 
\multicolumn{1}{c|}{} 
& \textbf{Ave} & \textbf{Gain} & \textbf{Ave} & \textbf{Gain} & \textbf{Ave} & \textbf{Gain} \\ \hline
\textbf{Equal weight} & 70.3 & - & 78.4 & - & 72.8 & - \\ \hline
\textbf{Design 1} & 68.8 & $-$1.5 & 79.5 & 1.1 & 72.5 & $-$0.3 \\ \hline
\textbf{Design 2} & 70.3 & 0 & 79.9 & 1.5 & \textbf{75.6} & \textbf{2.8} \\ \hline
\textbf{Design 3} & \textbf{70.9} & \textbf{0.6} & \textbf{81.0} & \textbf{2.6} & 73.2 & 0.4 \\ \hline\hline
\end{tabular}
}
\vspace{-5mm}
\end{table}

\MyParagraph{Datasets Preparation}
We follow emotion categorizations in psychology and communication literature to create a dataset for each categorization to evaluate the learnability.
The first dataset concerns the \emph{valence of emotion}\cite{Lazarus-1991}, including positive, negative, neutral as categorized in Table~\ref{tab:label_stat}.
The second dataset focuses on the \emph{resilient emotions} \cite{connor-2003}, including such \emph{individual-level resilience} emotions as \hope{} and \happiness, and such \emph{collective-level resilience} emotions as \empathy{} and \grateful.
The third dataset concerns the \emph{crisis responsibility attribution}\cite{jin-2014}, including \fear, \empathy, and \worry{} in the \emph{independent} category, and \anger, \sorrow, and \disappoint{} in the \emph{dependent} category.

\MyParagraph{Fine-tuned Neural Network as a Powerful Learning System}
We trained convolutional neural networks (CNN) as powerful learning systems for the tasks of classification for the three datasets.
Each input tweet is cleaned using standard NLP procedures to remove stop words and punctuation.
Each word is then converted to a vector of length 100 using Word2Vec with a window size of 5 trained on all available tweets in the dataset without removing low frequency words. 
Our CNN model, inspired from \cite{kim2014}, consists of 1-D convolutional masks of 3, 4, and 5 times of the feature length for each layer, respectively, 100 feature maps, and a dropout rate of 0.5.
The input of the CNN takes the first 50 words of a tweet, and zeros will be padded if the tweet is shorter than 50.

The split of the training, validation, and test sets is 60\%--20\%--20\%. 
To reduce the influence of the imbalanced dataset, we adopted a weight sampler to make the training set of equal size and used the balanced accuracy as the evaluation criterion for model selection at the validation stage. We selected the best model based on the lowest validation accuracy using an early stop of 10 epochs.
We trained one CNN for each combination of three datasets and four different voting weight designs (see Section~5.1).
Hyperparameters were selected via grid search separately on each dataset created using weight scheme \#2.
The learning rate and the batch size are 0.001 and 16 for the valence of emotion dataset, 0.003 and 16 for the resilient emotions dataset, and 0.03 and 32 for the crisis responsibility attribution dataset. 
For each combination, we repeated the training 150 times started from random data splitting, and reported the averaged balanced accuracy and F1-macro.
The performance of the learnability for different datasets is shown in the ``Equal Weight'' row of Table~\ref{tab:nn_results}.
It shows that the balanced accuracy can achieve 70.3\% to 78.4\% for different datasets.
This implies that the consensus by the majority voting on the annotated labels is learnable by powerful learning systems with accuracy greater than 70\%.

%% file: 5_discuss.tex
\section{Discussions and Experimental Design Recommendations}
\label{sec:dataset_mining}

\subsection{Weighted Voting Can Improve the Quality of Labels}
\begin{figure}[!t]
    \centering
    \includegraphics[width=0.49\linewidth, trim=0mm 0mm 12mm 0mm, clip]{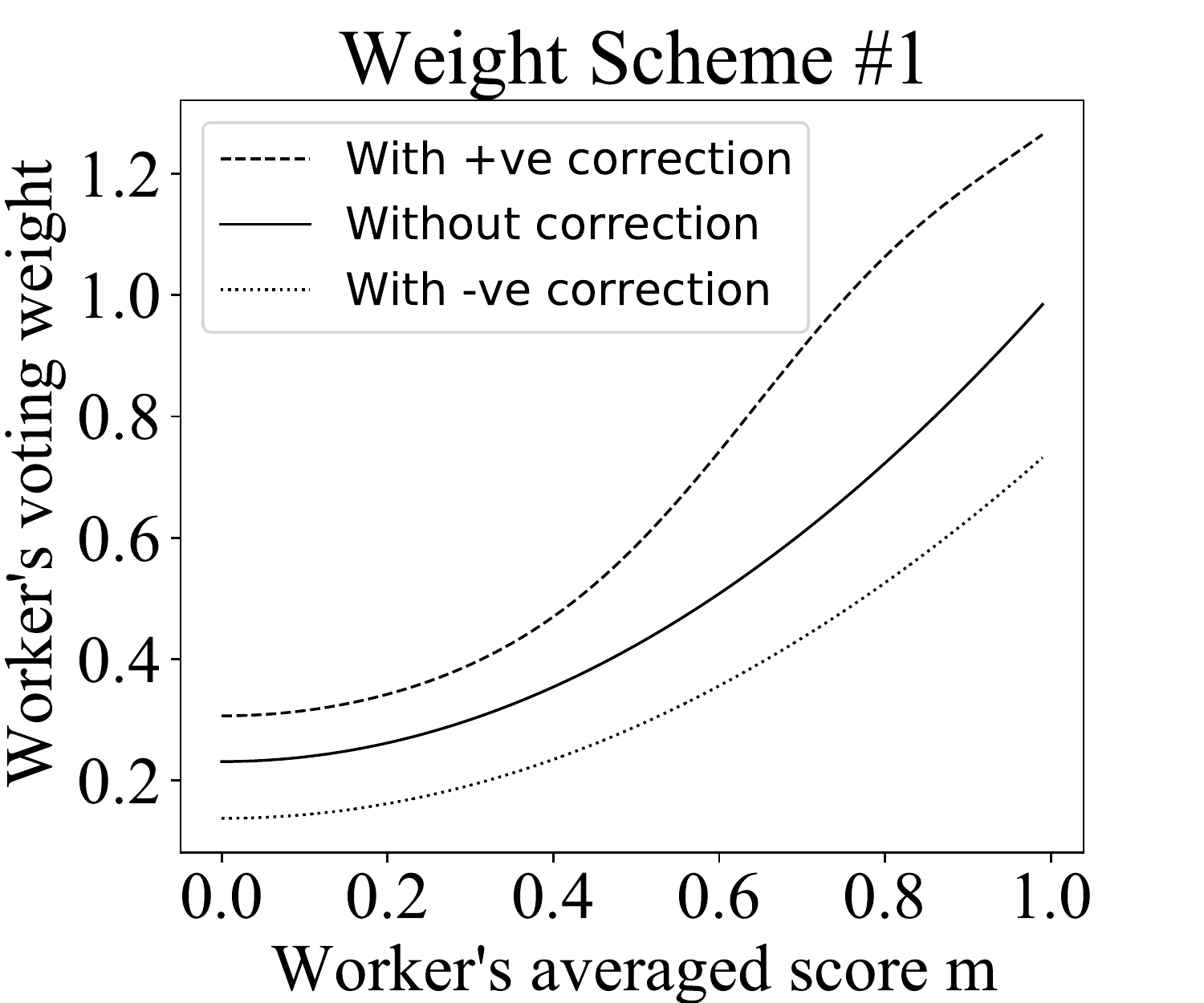} 
    \includegraphics[width=0.49\linewidth, trim=0mm 0mm 12mm 0mm, clip]{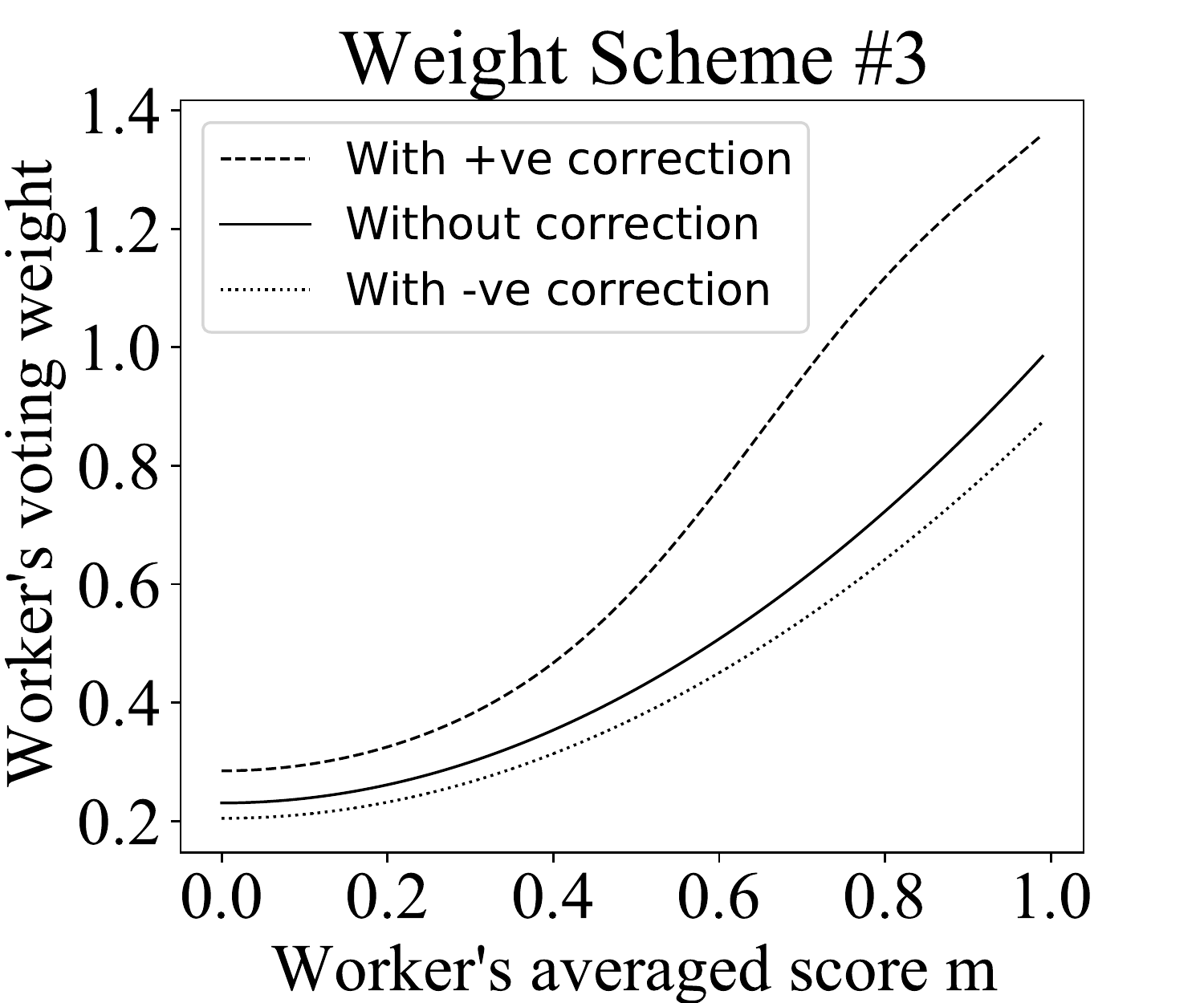} 
    \caption{Worker's voting weight scheme \#1 (left) and scheme \#3 (right). Higher averaged score $m$, smaller standard deviation $\sigma$, and larger slope $k$ leads to a higher voting weight.
    The top and bottom curves correspond to one-standard-derivation corrections.}
    \label{fig:weight_function}
\end{figure}

We investigate whether the majority voting results can be improved by assigning higher weights to good performing workers and/or lower weights to bad performing workers.
As the annotation system keeps track of a worker's historical quality scores that reflect his/her performance, we propose to use the following three features calculated from the score time series to quantify $i$th worker's performance.
First, the \emph{averaged score} or the cumulative quality score defined in Section~\ref{sec:system_design}, $m_i\in [0, 1]$, is used as the dominant feature to characterize the performance.
Second, the \emph{sample standard deviation} of the historical scores, $\sigma_i \in (0,\infty)$, represents the stability of the worker's performance.
Third, the upward/downward \emph{trend} or the slope of the score time series, $k_i \in \mathbb{R}$, is also be an informative indicator especially when there is a big drop in performance.
We will give a higher baseline voting weight to a worker with a higher averaged score, and will slightly boost the weight for a lower standard deviation and a positive trend.
We design three slightly different weight schemes according to the above principles. 
Weight schemes \#2 and \#3 will reward positive trend only but weight scheme \#1 will in addition penalize negative trend.
Weight scheme \#3 contains an extra boosting scalar for low-variance workers.
The design of the $\ell$th weight scheme, for $\ell \in \{1, 2\}$, is described as follows:
\begin{equation}
\begin{split}
w^{(\ell)}(m; \sigma, k) = \frac{m^2+a}{1+a} \bigg( 1 + \zeta_{\text{slope}}^{(\ell)} \, k/b \bigg) - 0.2\sigma \\
 - 2(\sigma-1) f \Big(5(m-0.8) \Big),
\end{split}
\end{equation}
where $m$, $\sigma$, and $k$ represent the averaged score, the standard deviation, and the slope of a worker; 
$\zeta_{\text{slope}}^{(\ell)} = \mathbbm{1}[k>0]$ for $\ell \in \{2, 3\}$ and $\zeta_{\text{slope}}^{(\ell)} = 1$ for $\ell=1$, where $\mathbbm{1}[\cdot]$ is the indicator function; 
and $f(x)$ is the kernel of the standard normal distribution.
Here, $w^{(3)}$ differs from $w^{(2)}$ by changing the additive correction term $-0.2\sigma$ into a multiplicative correction term $1-\sigma/c$.
Note that we use a quadratic function $\frac{m^2+a}{1+a}$ to transform the baseline score $m$ to reward excellent performing workers, and
all remaining scaling and additive terms involving $\sigma$ and $k$ are designed to lightly incorporate the worker's stability and performance trend.
We tuned hyperparameters to $a=0.3$, $b=2$, and $c=2$
by manually observing a reasonable assignment of the weight to each worker in 3-D space of $(m, \sigma, k)$.
Fig.~\ref{fig:weight_function} displays weight curves $w^{(\ell)}$ (black solid curves) against the averaged score $m$, with top and bottom curves showing positive and negative corrections in one standard derivation of $\{\sigma_i\}_{i=1}^n$ and $\{k_i\}_{i=1}^n$, where $n$ is the number of workers.

We tested the three variants of weighted-voting schemes with the same CNN structure but optimized individually.
The results on three datasets are shown in the last three rows of Table~\ref{tab:nn_results},
which reveals that weighted voting schemes \#2 and \#3 outperform the majority vote by 0.6\% to 2.8\% in terms of the balanced accuracy.
The worse performance of scheme \#1 implies that the negative impact of penalizing workers of decaying trend may outweigh the overall benefit brought by accounting for weights in voting.

\subsection{Challenges Due to Multiple-Emotion Tweets}
\label{subsec:multiple}
\begin{figure}[!t]
    \centering
    \includegraphics[width=0.495\linewidth, trim=2mm 0mm 20mm 0mm, clip]{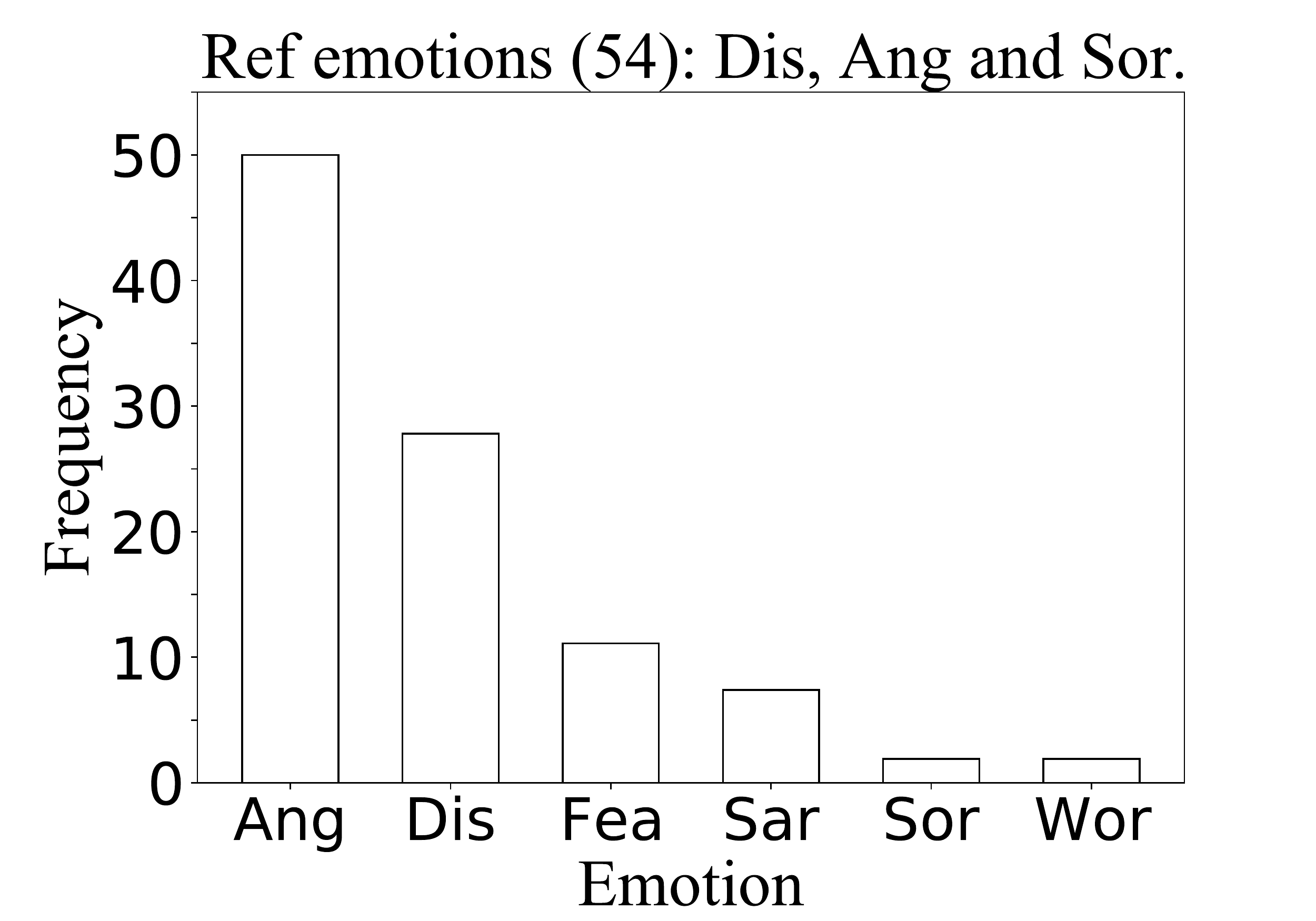} 
    \includegraphics[width=0.495\linewidth, trim=2mm 0mm 20mm 0mm, clip]{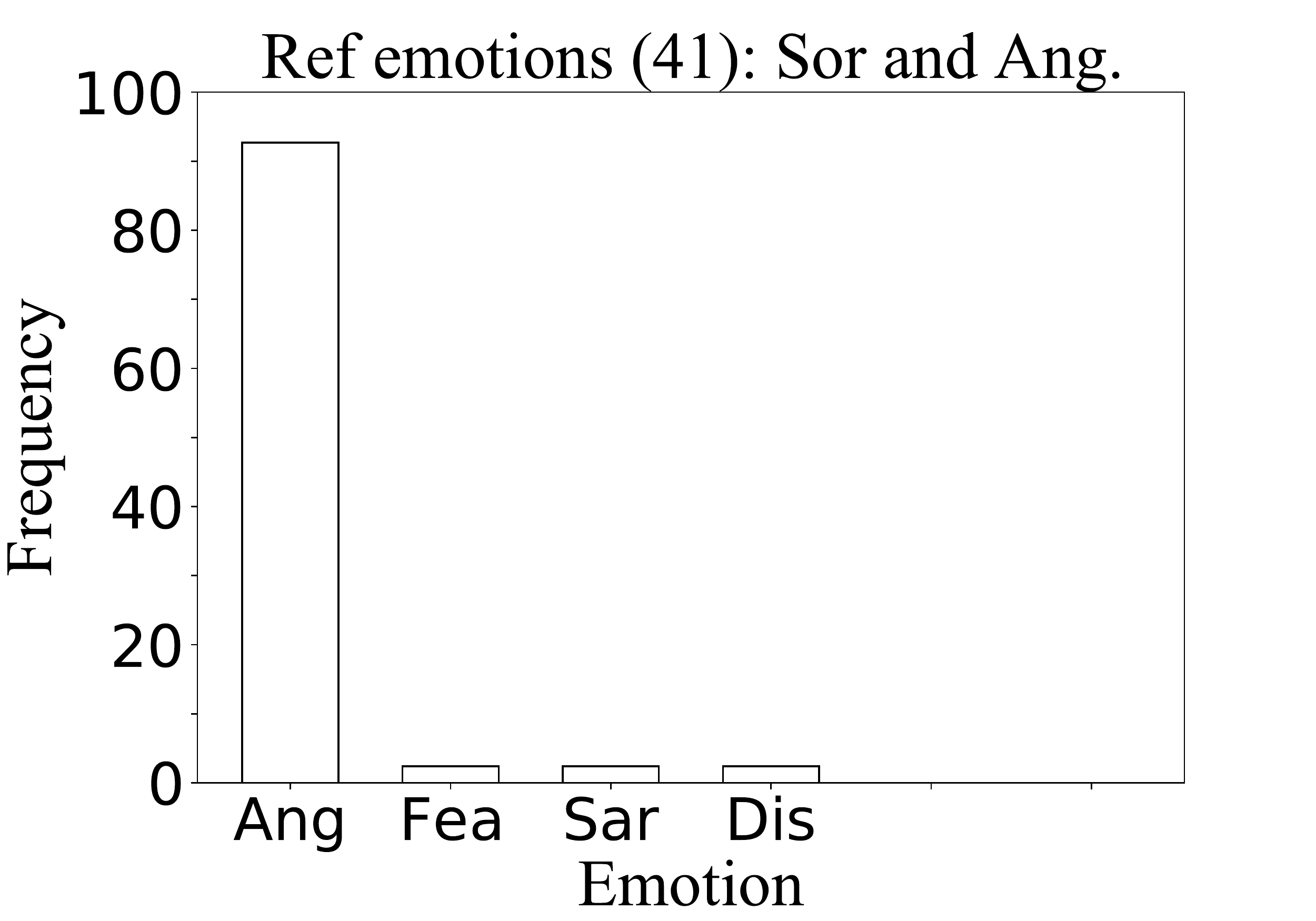} 
    \includegraphics[width=0.495\linewidth, trim=2mm 0mm 20mm 0mm, clip]{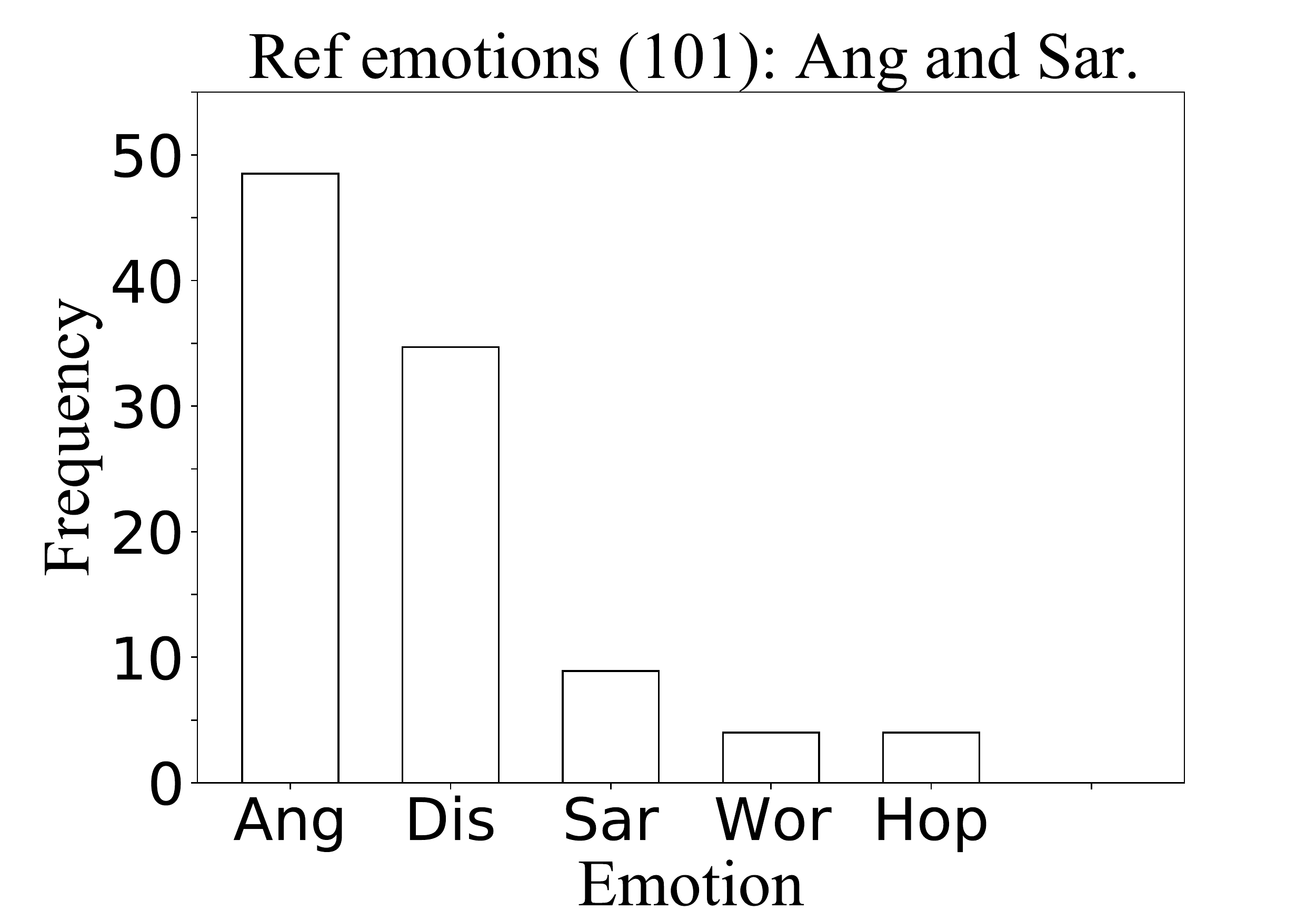} 
    \includegraphics[width=0.495\linewidth, trim=2mm 0mm 20mm 0mm, clip]{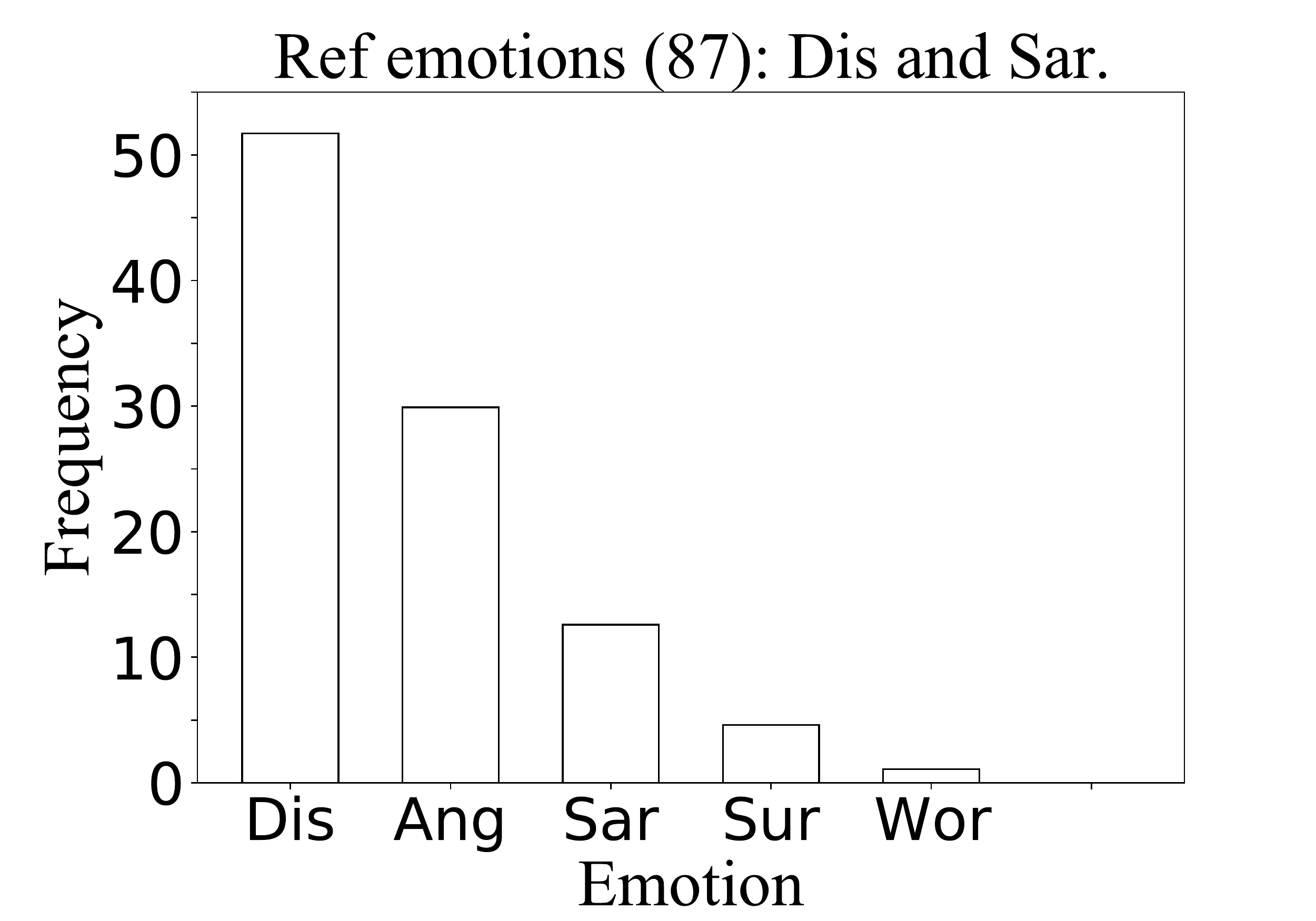} 
    \caption{Histograms of emotions for tweets submitted by workers. Each subplot corresponds to a single tweet with ground-truth labels shown in the title. In the first row, emotion \sorrow{} is masked by \anger{} and \disappoint{}. In the second row, emotion \sarcasm{} is masked by \anger{} and \disappoint{}.
    }
    \label{fig:specific_fig}
\end{figure}

Several patterns can be observed based on an inspection of the collected dataset.
First, we found that from workers' perceptions, more intuitive emotions tend to mask the less intuitive ones, based on an examination of the performance of workers on the expert-labeled tweets.
Fig.~\ref{fig:specific_fig} shows four examples, each of which displays the ground-truth labels and the histogram of emotions answered by workers. 
Figs.~\ref{fig:specific_fig}(a) and (b) show that emotion \sorrow{} is almost completely masked by \anger{} and \disappoint{} that usually have strong magnitude.
Figs.~\ref{fig:specific_fig}(c) and (d) show similar masking effect for emotion \sarcasm{}. 
Second, workers tend to just report the primary labels.
By examining the dataset, we found that 84$\%$ submitted answers contain one emotion.
In contrast, only 60\% of the tweets in the expert-labeled dataset have single labels.
This corner-cutting behavior can be explained by workers' motivation to boost the hourly earnings by reducing the time spent on difficult but less rewarding tasks.
Third, workers perform better when tweets have only one emotion.
The true positive rates for anger and disappoint are 71.9$\%$ and 71.3$\%$ for single-emotion tweets.
However, for tweets with both anger and disappoint emotions, the true positive rates are reduced to 50.3$\%$ and 49.7$\%$, respectively.

To make it more effective to accurately collect emotions especially for tweets with multiple emotions, 
we propose to use a crafted quality metric that encourages the discovery of secondary labels and improves the protocol to compensate for the extra time workers spend.
Alternatively, if a multiple-label task can be adapted into a single-label task with a reasonable amount of effort, doing so will avoid the hassle of crafting the quality metric while maintaining the accuracy of the collected labels.
The single-label task adaptation can save the time that workers assess the possibility of the existence of secondary labels and figure out what the labels are.

\subsection{Two Types of Misclassifications for Emotions}
\label{subsec:confusion}
\begin{figure}[!t]
    \centering
    \vspace{-3mm}
    \includegraphics[width=0.85\linewidth]{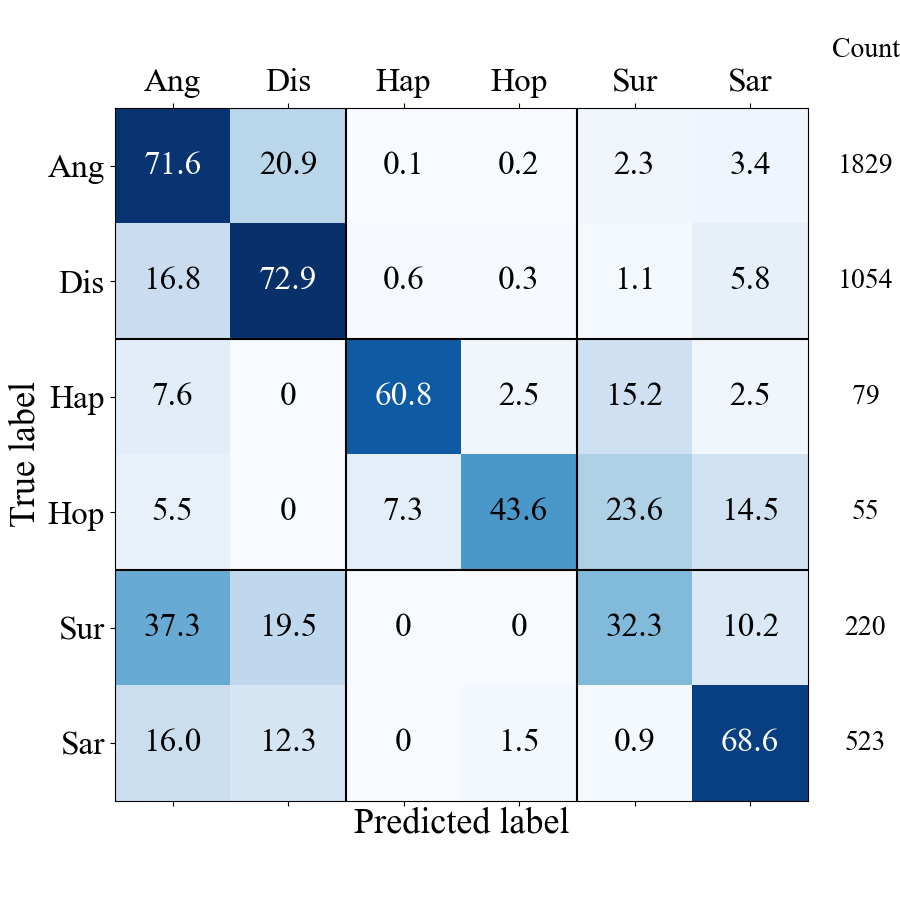} 
		\vspace{-7mm}
    \caption{Accuracy of annotated labels in terms of confusion matrix.}
		\vspace{-0mm}
    \label{fig:gov}
\end{figure}

We examine the accuracy of worker annotated labels and found that \anger{} and \disappoint{} are prone to false positive and \happiness{} and \hope{} are prone to false negative.
We evaluated the annotated labels on the expert labeled dataset and reported the classification accuracy in terms of the confusion matrix shown in Fig.~\ref{fig:gov}.
An accurate annotation will lead to a diagonal confusion matrix.
We examine the subset of tweets whose ground-truth emotion targets are the government, which is likely to reveal interesting patterns of this dataset since the local government of Flint was responsible for the water crisis.

The first major observation is that the false positives revealed by columns \anger{} and \disappoint{} of Fig.~\ref{fig:gov} may imply that workers reported their own emotions rather than those of tweets.
It is observed that 37.3\% of \surprise{} and 16.0\% \sarcasm{} were misclassified by workers as \anger{}.
Similarly, 19.5\% \surprise{} and 12.3\% \sarcasm{} are misclassified as \disappoint{}.
Note that both \surprise{} and \sarcasm{} are neutral emotions\cite{Lazarus-1991}, and we suspect the workers were unintentionally labeling their own emotional responses to the tweets, instead of labeling the emotions of the tweets.
Hence, it is important that, during worker's training session, practice questions should be designed to train workers from observers' standpoint, 
which will allow workers to assess more objectively the emotions rather than expressing their own viewpoints.
Workers should be reminded about this principle repeatedly throughout their duration of work to ensure that they work as expected.

The second major observation that is the false negatives revealed by rows \happiness{} and \hope{} of Fig.~\ref{fig:gov} may imply some tweets are more difficult than other.
Workers are confusing \happiness{} into \surprise{} at 15.2\% chance.
Workers are confusing \hope{} into \surprise{} at 23.6\% and \sarcasm{} at 14.5\% chance.
The statistics reveal an issue that the workers were not able to precisely comprehend the emotions of these tweets; and from researchers' perspective, these emotions are difficult to differentiate.
We examine the accuracy across individual tweets and found that it can vary drastically from 10$\%$ to 100$\%$.
According to the submitted labels for the expert labeled dataset, tweets with higher accuracy usually appears to have stronger magnitude of emotion with features such as using exclamation marks and containing such keywords as ``swearing'' for \anger{} and ``OMG'' for \surprise{}.
Tweets with lower accuracy are more often descriptive sentences or statements of neutral emotions, which may unintentionally trigger workers' own emotion about the tweet instead of faithfully answering the emotions of the tweet.
To increase the overall annotation accuracy, we propose to tailor the level of difficulty of a labeling question based on a worker's cumulative score.
The level of difficulty may be estimated by calculating the entropy\cite{cover2006elements} of the already submitted labels of the question. 
Questions with labels spread over various options are generally difficult ones and have a higher entropy, whereas those with concentrated labels are generally simple ones and have entropy close to zero.